%% file: main.tex
\documentclass[letterpaper]{article}

\input{tex/preamble}
\input{tex/metadata}

\begin{document}

\maketitle

\input{sections/00-abstract}
\input{sections/01-introduction}
\begin{figure*}[t!]
    \centering
    \includegraphics[width=0.96\textwidth]{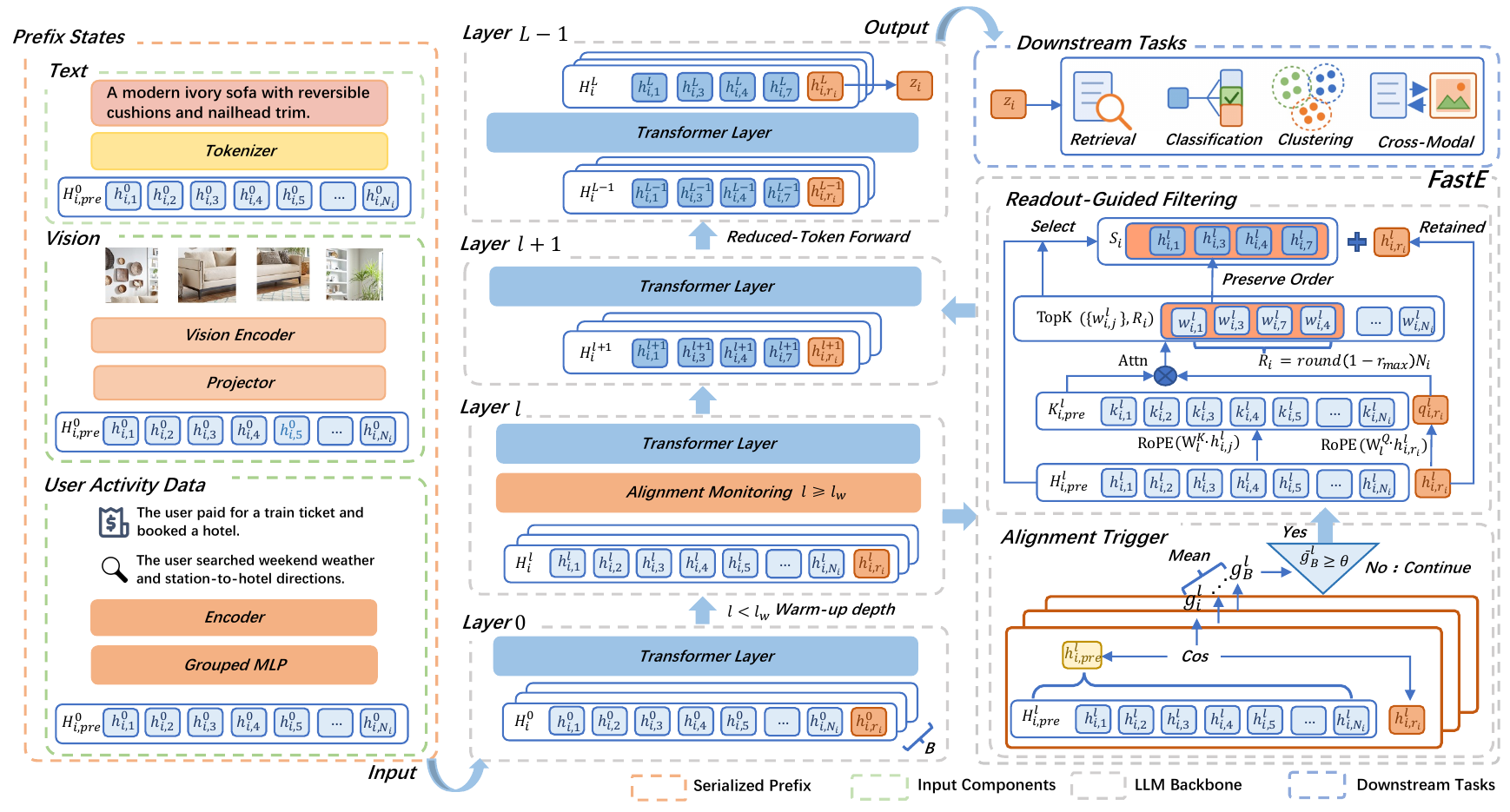}
    \caption{Overview of FastE for final-readout LLM embedding inference. FastE
preserves the full serialized prefix during warm-up and monitors batch-mean
readout--prefix alignment. At the first threshold crossing, readout-guided
attention-score ranking reduces the prefix to the target prefix-state budget;
subsequent transformer layers process the shortened sequence while preserving
the readout position and causal order, reducing their computation.}
\label{fig:faste-overview}
\end{figure*}
\input{sections/02-related-work}
\input{sections/03-method}

\input{sections/04-experiments}
\input{sections/05-limitations}
\input{sections/06-conclusion}

\clearpage
\IfFileExists{references.bib}{\bibliography{references}}{}

\clearpage
\appendix
\setcounter{secnumdepth}{2}

\input{appendices/01-additional-experimental-results}

\end{document}

%% file: tex/preamble.tex
\usepackage[preprint]{aaai2027}  
\usepackage[hyphens]{url}  
\usepackage{graphicx} 
\usepackage{natbib}  
\usepackage{caption} 
\long\def\zl#1{%
\noindent{\color{brown}#1}%
}

\usepackage{amsmath}
\usepackage{amssymb}
\usepackage{booktabs}
\usepackage{multirow}
\usepackage{hhline}
\usepackage{algorithm}
\usepackage{algpseudocode}
\usepackage[table]{xcolor}
\usepackage{dblfloatfix}

%% file: tex/metadata.tex
\title{FastE: Readout-Triggered Token Compression for LLM Embedding Inference}

\author{
Jinsong Shu\textsuperscript{\rm 1,4},
Jinyong Wen\textsuperscript{\rm 2},
Baokun Wang\textsuperscript{\rm 2},
Zhongle Xie\textsuperscript{\rm 1}\thanks{Corresponding author.}\\
Lidan Shou\textsuperscript{\rm 3,4},
Weiqiang Wang\textsuperscript{\rm 2},
Gang Chen\textsuperscript{\rm 1}
}

\affiliations{
\textsuperscript{\rm 1}Zhejiang University
\qquad
\textsuperscript{\rm 2}Ant Group\\
\textsuperscript{\rm 3}The State Key Laboratory of Blockchain and Data Security,
Zhejiang University\\
\textsuperscript{\rm 4}Hangzhou High-Tech Zone (Binjiang) Institute of
Blockchain and Data Security
}

%% file: sections/00-abstract.tex
\begin{abstract}
In this study, we identify depth-dependent prefix redundancy in final-readout
LLM embedding models, notably across representative backbones including
Qwen3-Embedding and Qwen3-VL-Embedding.
We find that removing prefix states is substantially more damaging in shallow
layers than at greater depth, showing that prefix states become increasingly
compressible as the prefix and readout states propagate through the network.
To this end, we introduce FastE, a training-free, plug-and-play
method. FastE uses a shared fixed threshold on batch-mean readout--prefix
alignment as a lightweight online heuristic for selecting when compression occurs, and
ranks prefix states by the attention scores they receive from the readout position
to determine which states are retained in subsequent layers.
Our evaluations demonstrate FastE's ability to substantially reduce
computational costs: on NarrativeQA with Qwen3-Embedding-0.6B, it reduces
decoder-backbone FLOPs by 40.11\% while retaining 99.53\% of Full Forward
nDCG@10.
Across five text embedding benchmarks, two backbone scales, and three cross-modal
retrieval tasks, the quality--efficiency trade-off is directly customizable
through the maximum removal ratio without retraining.
We believe FastE offers practical value for scalable embedding generation
in retrieval, indexing, clustering, and multimodal representation systems.
\end{abstract}

%% file: sections/01-introduction.tex
\section{Introduction}

\begin{figure*}[t!]
    \centering
    \begin{minipage}[b]{0.62\textwidth}
        \centering
        \includegraphics[width=\linewidth]{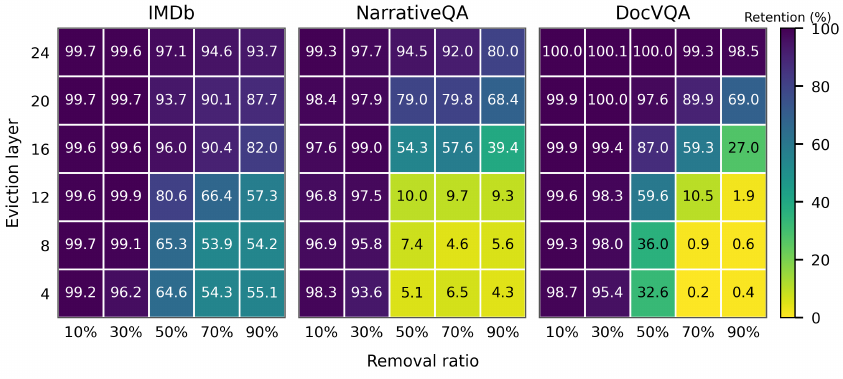}\\[-0.5ex]
        {\small (a) Performance retention}
    \end{minipage}\hspace{0.006\textwidth}
    \begin{minipage}[b]{0.27\textwidth}
        \centering
        \includegraphics[width=\linewidth]{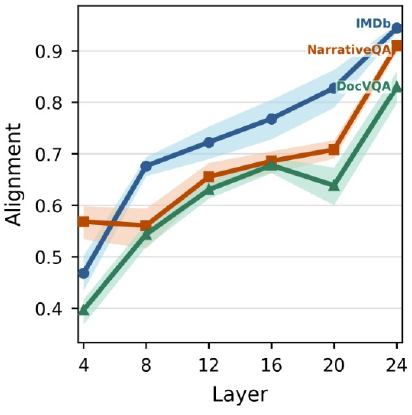}
        \\[-0.5ex]
        {\small (b) Readout--prefix alignment}
    \end{minipage}
    \caption{Depth-wise prefix compressibility in final-readout LLM embedding
    inference. (a) Performance retention across removal ratios and eviction layers;
    later eviction better preserves quality under aggressive removal. (b)
    Readout--prefix alignment across depth for three task--backbone settings.}
    \label{fig:observation}
\end{figure*}

Embeddings are vector representations that support retrieval, indexing, classification, and clustering across text and multimodal applications~\citep{10.1145/3617833,opitz-etal-2025-interpretable}.
Large language model (LLM)-based embedding models have recently become strong general-purpose embedding backbones by transferring the semantic capabilities of pretrained LLMs to embedding tasks for sequence inputs~\citep{zhang2025qwen3embedding,lee2025nvembed,wang-etal-2024-improving-text,behnamghader2024llmvec,muennighoff2025generative}.

We focus on final-readout LLM embedding models, which serialize each input as a prefix followed by an end-of-sequence (EOS) or dedicated readout token and process it with causal attention.
The hidden state at this final position, namely \textit{readout state}, develops layer by layer and ultimately yields the sequence embedding.
The preceding serialized input forms the \textit{serialized prefix} and its layer-wise hidden states, which are called \textit{prefix states}, provide the context absorbed by the readout state.
Note that standard inference propagates both through the transformer stack.

{
For long serialized prefixes, this joint propagation becomes computationally expensive.
In an Alipay production workload, processing 450 million inputs with a mean valid serialized-prefix length of 732 tokens using Q-Anchor Embedding~\citep{yuan2026queryanchor}, an industrial user-representation model developed at Ant Group, requires approximately 6 hours on 300 L20 GPUs.
A natural response is prefix-state filtering, which shortens subsequent computation by removing selected prefix states.
Attention-based pruning and token merging provide such reduction operators, but commonly apply them at fixed depths or rely on modality-specific signals~\citep{chen2024fastv,zhang2025sparsevlm,Yang_2025_CVPR,10.1007/978-3-031-72643-9_13}.
Key--value (KV) cache compression instead selects past states for future autoregressive queries~\citep{xiao2023streamingllm,li2024snapkv}, yet its decoding objective does not characterize how prefix states support a single readout state as it develops through depth.
Hence, these limitations leave a central question: \textit{are prefix states equally necessary in shallow and deep layers?}
}

{
Our controlled depth-wise interventions reveal a clear answer: no, \textit{prefix-state removal causes substantially less performance loss in deeper layers}.
We call this empirical pattern \textbf{depth-dependent prefix redundancy}, in which prefix states that are important during shallow computation become more compressible after further contextualization.
As Figure~\ref{fig:observation}(a) quantifies: at 50\% prefix-state removal, the performance drop decreases from 35.4--94.9\% at layer 4 to 0--5.5\% at layer 24 across the three evaluated settings.
Furthermore, the depth at which removal incurs limited performance loss varies across task--backbone settings, making a single fixed trigger layer less reliable across settings.
To identify an online signal for this depth effect, Figure~\ref{fig:observation}(b) traces readout--prefix alignment, the cosine similarity between the readout state and the mean prefix-state representation, across depth.
The alignment trajectories generally increase with depth, suggesting a lightweight heuristic for identifying when prefix-state compression becomes better tolerated without treating alignment as causal proof of redundancy.
}

{
Guided by these findings, we introduce FastE, a training-free method that decomposes prefix-state compression into two decisions: \textit{when} compression should begin and \textit{which} prefix states should be retained.
To decide \textit{when} compression should begin, FastE preserves the full serialized prefix during an initial warm-up and uses a fixed threshold on batch-mean readout--prefix alignment, shared across the reported text tasks, to select the online trigger layer.
To decide \textit{which} prefix states should be retained, FastE applies readout-guided attention-score ranking at the first threshold crossing and performs one-shot prefix compression to the configured prefix-state budget while preserving the readout position, causal order, and original position identifiers.
Across five text tasks and two Qwen3-Embedding scales, FastE outperforms transferred baselines under matched prefix-state budgets, while Qwen3-VL-Embedding and E5-Mistral results provide initial evidence across modalities and model architectures.
On NarrativeQA with Qwen3-Embedding-0.6B, FastE retains 99.53\% of Full Forward nDCG@10 while reducing decoder-backbone FLOPs by 40.11\% and achieving a 1.363$\times$ measured end-to-end speedup.
Together, the depth-wise finding, the readout-guided \textit{when}/\textit{which} design, and the measured quality--efficiency evaluation constitute the paper's main contributions.
}

%% file: sections/02-related-work.tex
\section{Related Work}

\noindent
\textbf{Final-readout LLM embedding models.}
{
Final-readout LLM embedding models derive the sequence embedding from the final-layer state of an EOS or dedicated readout token, whose causal attention covers the serialized prefix~\citep{wang2024improving,zhang2025qwen3embedding}
Post-generation mean pooling can improve representations in general-purpose language models~\citep{wang2026truth}, while trainable final-layer pooling provides another aggregation architecture~\citep{tang2024pooling}.
FastE instead accelerates representation construction in final-readout LLM embedding models by filtering prefix states and does not assume transfer to pooling architectures.
}

\noindent
\textbf{Autoregressive inference and KV-cache compression.}
{
Long-context generation methods retain attention sinks, heavy hitters, selected prompt tokens, or compressed KV entries to reduce repeated decoding after prompt encoding~\citep{xiao2023streamingllm,li2024snapkv}; HybridKV extends this regime to multimodal generation~\citep{zeng2026hybridkv}.
FastE addresses a different regime: embedding inference performs one forward pass to produce one sequence embedding, and prefix-state filtering reduces attention and feed-forward (MLP) computation in subsequent transformer blocks.
}

\noindent
\textbf{Sequence reduction during representation construction.}
{
For encoder-only sentence embedding, attention-based pruning aggregates importance across query positions and removes tokens at fixed layers~\citep{qi2024practical}.
Multimodal reduction methods use fixed or progressive schedules, visual or cross-modal signals, pre-LLM selection, adaptive information flow, or token merging~\citep{chen2024fastv,zhang2025sparsevlm,yang2025visionzip,wan2026rtprune,tong2025flowcut,xing2025pyramiddrop,bolya2023tome,shang2025llavaprumerge,alvar2025divprune,jiang2025dcp,ye2025fitprune}.
FastE differs by using the evolving readout state to determine both the trigger layer and the retained prefix states, without relying on visual structure or another modality-specific signal.
}

%% file: sections/03-method.tex
\section{Method}

\subsection{Problem Formulation}

We consider a final-readout LLM embedding model that processes serialized inputs
with causal attention. Given a sample $x_i$ containing one or more modalities, the
model first maps each input component to an initial sequence of vectors. A
tokenizer and the model's token embedding layer process text, a pretrained
vision encoder and its projector process images, and structured inputs are
serialized into the same embedding space. The resulting vectors, together with
textual instructions and special tokens, are concatenated into a unified
sequence and processed by the LLM backbone~\citep{li2026qwen3vlembedding}. FastE operates on
this unified hidden-state sequence after serialization.

For sample $i$, let $r_i$ denote the position of the readout token, whose
final-layer hidden state is normalized to obtain the sequence embedding.
Depending on the backbone, the readout token may be the model's existing
sequence-ending token (e.g., EOS) or a dedicated token introduced for embedding
extraction. The $N_i$ positions preceding $r_i$ form the serialized prefix,
and the forward pass is
\begin{equation}
    \begin{aligned}
    H_i^0
    &= [h_{i,1}^0,\ldots,h_{i,N_i}^0;h_{i,r_i}^0],\\
    H_i^{l+1}
    &= F_l(H_i^l),\quad 0\leq l<L,\\
    z_i
    &= \operatorname{Norm}\!\left(h_{i,r_i}^{L}\right).
    \end{aligned}
    \label{eq:representation-forward}
\end{equation}
We distinguish sequence positions from their layer-wise representations:
$h_{i,j}^l$ denotes the hidden state at position $j$ in layer $l$. Before
filtering at layer $l$, $H_{i,\mathrm{pre}}^l=[h_{i,1}^l,\ldots,h_{i,N_i}^l]$
is the serialized prefix. Every prefix state is considered for compression,
regardless of modality; the readout token is always preserved. Removed prefix states and their
corresponding positions no longer participate in the current or subsequent
transformer layers.

\subsection{Depth-Wise Prefix Compressibility}
\label{sec:depth-redundancy}

We first examine whether the effect of prefix removal depends on the layer at
which removal is applied. To isolate this factor, we conduct a controlled
inference-time intervention following the diagnostic protocol for depth-wise
token compressibility~\citep{xing2025pyramiddrop}. We vary two factors: the
layer at which eviction occurs and the fraction of prefix states
removed. For a serialized prefix of length $N$ and removal ratio $r$, we set
the retained count to $K=\max(1,\operatorname{round}((1-r)N))$, partition the
prefix into $K$ contiguous, nearly equal-sized blocks, and retain only the
causally last state in each block. The same deterministic position-based rule
is used at every tested layer, without attention or learned scores,
to isolate compression depth from state-ranking quality. Each run performs a
single removal event, excluding the effect of predefined multi-stage schedules,
and all subsequent transformer blocks run unchanged. We perform this
intervention on IMDb, NarrativeQA, and DocVQA so that the observation spans
classification, long-document retrieval, and visual-document retrieval.

Figure~\ref{fig:observation}(a) shows that removing a matched fraction of
prefix states causes substantially greater quality loss in shallow
layers than in deeper layers, especially under aggressive removal. However, the
transition depth varies across task--backbone settings, indicating that a
predefined global compression layer may not transfer reliably across settings.
This motivates an online signal for selecting the trigger layer.

To obtain a lightweight online signal, we monitor the readout state. In the
final-readout LLM embedding models considered here, its final-layer state forms
the sequence embedding and receives sequence-level contrastive
supervision~\citep{zhang2025qwen3embedding}, while causal attention allows the
readout position to access the complete serialized prefix. We therefore use
readout--prefix alignment to characterize its relation to the current prefix.
For each sample,
we denote the mean prefix representation by $\bar h_{i,\mathrm{pre}}^l$ and
define its alignment with the readout state as
\begin{equation}
    \bar h_{i,\mathrm{pre}}^l =
    \operatorname{Mean}(H_{i,\mathrm{pre}}^l),
    \qquad
    g_i^l = \cos\!\left(h_{i,r_i}^l,\bar h_{i,\mathrm{pre}}^l\right).
    \label{eq:sample-readiness}
\end{equation}

Figure~\ref{fig:observation}(b) shows that mean readout--prefix alignment
generally increases with depth, while the three task--backbone settings follow
different trajectories. The trend is consistent with the depth-wise tolerance
in Figure~\ref{fig:observation}(a). FastE therefore uses batch-mean alignment as
a lightweight heuristic for choosing when to compress, as detailed below.

\subsection{FastE}

\paragraph{Overview.}

Based on these observations, we propose FastE, a readout-triggered method
for representation compression that exploits depth-dependent prefix
compressibility in embedding models. To avoid information loss from
premature compression in shallow layers, FastE first preserves the full
prefix and then monitors the alignment between the readout state and
the prefix states in subsequent layers. Once the alignment score reaches a
threshold, FastE ranks prefix states by the attention scores they receive from
the readout position and immediately reduces the sequence to the target budget. All
subsequent transformer layers operate on this shortened sequence without
additional eviction decisions. Figure~\ref{fig:faste-overview} summarizes this
inference path across serialized text, visual, and structured inputs.
Algorithm~\ref{alg:faste} in the appendix gives the
complete inference procedure.

The procedure is controlled by warm-up depth $l_{\mathrm w}$, alignment
threshold $\theta$, and maximum removal ratio $r_{\max}$. Since filtering
considers all $N_i$ serialized-prefix states, the final target for sample $i$ is
\begin{equation}
    R_i
    = \max\!\left(
       1,\operatorname{round}\!\left((1-r_{\max})N_i\right)
       \right).
    \label{eq:final-budget}
\end{equation}
The readout position is excluded from both the budget and Top-K competition.

\paragraph{Readout-triggered start depth.}

FastE uses readout--prefix alignment to determine when compression begins.
After the warm-up, it evaluates the sample-level score $g_i^l$ from
Eq.~\eqref{eq:sample-readiness} at every layer $l_{\mathrm w}\leq l<L$.
For the current inference batch $B$, we average the sample scores to obtain a
batch-level alignment signal for deciding whether the entire batch should
trigger compression:
\begin{equation}
    \bar g_B^l =
    \frac{1}{|B|}\sum_{i\in B}g_i^l,
    \label{eq:batch-readiness}
\end{equation}
\begin{equation}
    l_B =
    \min\!\left(
    \left\{
    l:\,l_{\mathrm w}\leq l<L,
    \bar g_B^l\geq\theta
    \right\}
    \cup\{L-1\}
    \right).
    \label{eq:threshold-trigger}
\end{equation}
Thus, $l_B$ is the first layer whose batch-mean alignment reaches $\theta$.
If no earlier layer crosses the threshold, the union selects $L-1$ and applies
the same filtering operation before the final transformer block. The
final-layer fallback ensures a defined compression path when a batch never
crosses the threshold, with limited savings as only one block remains.
\paragraph{Readout-guided prefix-state filtering.}

The alignment score determines only when compression occurs. At the input to
the trigger layer $l_B$, FastE filters the complete budget
$N_i-R_i$ at the selected layer. It computes the attention score that each
prefix state receives from the readout position using the trigger layer's
current Q/K projection weights. To keep the notation compact,
we suppress only the attention-head index. Let $h_{i,r_i}^{l_B}$ be the readout
state and $h_{i,j}^{l_B}$ the prefix state at position $j$. Using the original
rotary position associated with each state gives
\begin{equation}
    q_{i,r_i}^{l_B} =
    \operatorname{RoPE}\!\left(
    W_{l_B}^Q h_{i,r_i}^{l_B}\right),
    \qquad
    k_{i,j}^{l_B} =
    \operatorname{RoPE}\!\left(
    W_{l_B}^K h_{i,j}^{l_B}\right).
    \label{eq:readout-qk}
\end{equation}
The importance score of the prefix state at position $j$ is
\begin{equation}
    w_{i,j}^{l_B} = \operatorname{AvgHeads}\!\left[
    \operatorname{softmax}\!\left(
    \frac{(q_{i,r_i}^{l_B})^\top k_{i,j}^{l_B}}{\sqrt{d_h}}
    \right)\right],
    \label{eq:readout-importance}
\end{equation}
For an attention layer with $A$ query heads and $A_{\mathrm{kv}}$ key--value
heads, where $A_{\mathrm{kv}}\mid A$, query head $h$ is paired with key--value
head $\lfloor hA_{\mathrm{kv}}/A\rfloor$. This standard mapping covers MHA
($A_{\mathrm{kv}}=A$), GQA ($1<A_{\mathrm{kv}}<A$), and MQA
($A_{\mathrm{kv}}=1$); the softmax is taken over active prefix positions per
query head before $\operatorname{AvgHeads}$ averages the resulting $A$
distributions. Here, $d_h$ is the head dimension. The readout position is
excluded. Let
$\mathbf w_i^{l_B}$ collect the scores over all prefix positions.
FastE retains the $R_i$ highest-scoring prefix states:
\begin{equation}
    \mathcal{S}_i =
    \operatorname{TopK}\!\left(\mathbf w_i^{l_B},R_i\right).
    \label{eq:readout-topk}
\end{equation}
The surviving prefix states retain their original causal order and rotary
position IDs, while the readout state remains at the final sequence position
and produces the embedding through the original extraction rule. The trigger
layer and all subsequent layers then operate on $R_i+1$ hidden states.

\paragraph{Efficiency analysis.}

FastE shortens each sequence from $N_i+1$ to $R_i+1$ before
the trigger layer $l_B$, so all remaining transformer blocks
process fewer hidden states. We report the realized decoder-backbone
FLOPs reduction as
\begin{equation}
    \rho_{\mathrm{FLOPs}}
    =
    1-
    \frac{
    \sum_{l=0}^{L-1}
    \mathcal{C}_l(T_l^{\mathrm{FastE}},P_l^{\mathrm{FastE}})
    }{
    \sum_{l=0}^{L-1}
    \mathcal{C}_l(T_l^{\mathrm{Full}},P_l^{\mathrm{Full}})
    }.
    \label{eq:flops}
\end{equation}
Here, $\mathcal{C}_l(T_l,P_l)$ denotes the FLOPs of the projection,
attention, and MLP operations in transformer layer $l$, while $T_l$
and $P_l$ are the realized padded-token and attention-pair counts.
GQA-aware formulas and the shape-pass protocol are in the appendix.
Equation~\eqref{eq:flops} reports backbone savings only: alignment costs
$O(N_iH)$ per monitored layer, and triggering additionally requires worst-case
$O(N_iH^2)$ readout-Q/prefix-K projections, one $O(N_iH)$ attention row, and a
one-shot Top-K without a full $N_i\times N_i$ matrix. These controller costs
are excluded from the FLOPs reduction but included in measured GPU-forward and
E2E time, then amortized over the shortened remaining layers. Unlike FastV's
fixed-layer all-query visual pruning and ToMe's state merging, FastE uses the
readout state to determine both when and what to compress
\citep{chen2024fastv,bolya2023tome}: it waits for sufficient readout--prefix
alignment, then retains states with current-layer readout attention while
preserving causal order and rotary positions.

\input{sections/04-main-tables}

%% file: sections/04-main-tables.tex
\begin{table*}[!t]
\centering
\begin{minipage}[t]{0.69\textwidth}
\vspace{0pt}\centering\tiny
\setlength{\tabcolsep}{1.8pt}
\renewcommand{\arraystretch}{1.09}
\caption{Text-embedding performance under matched per-sample target prefix-state
budgets. Mean retention averages task-wise scores relative to Full Forward.
Boldface marks the best compressed result for each metric and removal ratio.}
\label{tab:main}
\resizebox{\linewidth}{!}{%
\begin{tabular}{@{}c|l|cccccc@{}}
\hhline{-|-|------}
Model & Method & NQA & Core17 & Supply & IMDb & ArXiv & Retention \\
 & & nDCG $\uparrow$ & nDCG $\uparrow$ & mAP $\uparrow$ & Acc. $\uparrow$ & V-meas. $\uparrow$ & (\%) $\uparrow$ \\
\hhline{-|-|------}
\multirow{13}{*}{\rotatebox[origin=c]{90}{\textit{Qwen3-Embedding-0.6B}}}
 & Full Forward (100\%) & 0.4541 & 0.4314 & 0.8700 & 0.9316 & 0.5054 & 100.00 \\
\hhline{~|-|------}
 & \multicolumn{1}{c|}{\cellcolor{gray!15}} & \multicolumn{6}{c}{\cellcolor{gray!15}\textit{30\% Prefix-State Removal ($r_{\max}=0.30$)}} \\
 & ToMe (ICLR'23) & 0.4015 & 0.3916 & 0.8638 & 0.9223 & 0.5023 & 95.37 \\
 & FastV (ECCV'24) & 0.4363 & 0.4052 & \textbf{0.8720} & 0.8992 & 0.4979 & 97.05 \\
 & RTPrune (ICML'26) & 0.3856 & \textbf{0.4398} & 0.8719 & 0.8990 & \textbf{0.5083} & 96.83 \\
 & OptScale (ICML'26) & 0.4522 & 0.3850 & 0.8664 & 0.9223 & 0.5050 & 97.47 \\
 & \cellcolor{green!10}FastE (Ours) & \cellcolor{green!10}\textbf{0.4600} & \cellcolor{green!10}0.4293 & \cellcolor{green!10}0.8691 & \cellcolor{green!10}\textbf{0.9270} & \cellcolor{green!10}0.5039 & \cellcolor{green!10}\textbf{99.98} \\
\hhline{~|-|------}
 & \multicolumn{1}{c|}{\cellcolor{gray!15}} & \multicolumn{6}{c}{\cellcolor{gray!15}\textit{70\% Prefix-State Removal ($r_{\max}=0.70$)}} \\
 & ToMe (ICLR'23) & 0.3711 & 0.3138 & 0.8450 & \textbf{0.9027} & 0.3981 & 85.45 \\
 & FastV (ECCV'24) & 0.3437 & 0.3175 & 0.8492 & 0.7902 & 0.4624 & 84.64 \\
 & RTPrune (ICML'26) & 0.1690 & 0.3307 & 0.8269 & 0.7720 & 0.4292 & 75.34 \\
 & OptScale (ICML'26) & 0.2912 & 0.1873 & \textbf{0.8662} & 0.8921 & 0.4661 & 79.02 \\
 & \cellcolor{green!10}FastE (Ours) & \cellcolor{green!10}\textbf{0.4318} & \cellcolor{green!10}\textbf{0.4032} & \cellcolor{green!10}0.8654 & \cellcolor{green!10}0.8974 & \cellcolor{green!10}\textbf{0.4797} & \cellcolor{green!10}\textbf{95.85} \\
\hhline{-|-|------}
\multirow{13}{*}{\rotatebox[origin=c]{90}{\textit{Qwen3-Embedding-4B}}}
 & Full Forward (100\%) & 0.5824 & 0.5211 & 0.8953 & 0.9492 & 0.5158 & 100.00 \\
\hhline{~|-|------}
 & \multicolumn{1}{c|}{\cellcolor{gray!15}} & \multicolumn{6}{c}{\cellcolor{gray!15}\textit{30\% Prefix-State Removal ($r_{\max}=0.30$)}} \\
 & ToMe (ICLR'23) & 0.5190 & 0.4634 & 0.8825 & 0.9179 & 0.4314 & 91.39 \\
 & FastV (ECCV'24) & 0.5565 & 0.5015 & 0.8954 & 0.9297 & 0.5079 & 97.64 \\
 & RTPrune (ICML'26) & 0.5405 & 0.5009 & 0.8921 & 0.9312 & 0.5126 & 97.21 \\
 & OptScale (ICML'26) & 0.5679 & 0.4689 & 0.8963 & 0.9116 & 0.4418 & 93.86 \\
 & \cellcolor{green!10}FastE (Ours) & \cellcolor{green!10}\textbf{0.5800} & \cellcolor{green!10}\textbf{0.5051} & \cellcolor{green!10}\textbf{0.8971} & \cellcolor{green!10}\textbf{0.9502} & \cellcolor{green!10}\textbf{0.5140} & \cellcolor{green!10}\textbf{99.29} \\
\hhline{~|-|------}
 & \multicolumn{1}{c|}{\cellcolor{gray!15}} & \multicolumn{6}{c}{\cellcolor{gray!15}\textit{70\% Prefix-State Removal ($r_{\max}=0.70$)}} \\
 & ToMe (ICLR'23) & 0.4801 & 0.3988 & 0.8567 & 0.8845 & 0.3797 & 84.29 \\
 & FastV (ECCV'24) & 0.4663 & 0.4527 & 0.8746 & 0.8454 & 0.4952 & 89.94 \\
 & RTPrune (ICML'26) & 0.3613 & 0.3662 & 0.8819 & 0.8286 & 0.4836 & 82.37 \\
 & OptScale (ICML'26) & 0.4230 & 0.3252 & 0.8565 & 0.8542 & 0.4449 & 81.39 \\
 & \cellcolor{green!10}FastE (Ours) & \cellcolor{green!10}\textbf{0.5464} & \cellcolor{green!10}\textbf{0.4959} & \cellcolor{green!10}\textbf{0.8953} & \cellcolor{green!10}\textbf{0.9456} & \cellcolor{green!10}\textbf{0.4955} & \cellcolor{green!10}\textbf{96.93} \\
\hhline{-|-|------}
\end{tabular}}
\end{minipage}\hfill
\begin{minipage}[t]{0.30\textwidth}
\vspace{0pt}\centering\footnotesize
\setlength{\tabcolsep}{1.2pt}\renewcommand{\arraystretch}{1.17}
\caption{Representative AUC on Alipay industrial prediction tasks at 30\% and 70\% removal.
All-39 avg. covers all 39 tasks.}
\label{tab:industrial-selected}
\begin{tabular*}{\linewidth}{@{\extracolsep{\fill}}lccc@{}}
\toprule
\multirow{2}{*}{Task} & \multirow{2}{*}{Full} & \multicolumn{2}{c}{FastE} \\
\cmidrule(lr){3-4} & & 30\% & 70\% \\
\midrule
User access & 0.8413 & 0.8413 & 0.8347 \\
Click label & 0.6350 & 0.6353 & 0.6348 \\
User churn & 0.9703 & 0.9703 & 0.9683 \\
Forest activity & 0.9587 & 0.9587 & 0.9558 \\
Game activity & 0.8985 & 0.8986 & 0.8807 \\
Game purchase & 0.9459 & 0.9458 & 0.9390 \\
Sim. game & 0.9495 & 0.9493 & 0.9467 \\
Offline payment & 0.7645 & 0.7643 & 0.7591 \\
Reward use & 0.7373 & 0.7371 & 0.7350 \\
Payment click & 0.8903 & 0.8909 & 0.8880 \\
Card status & 0.9899 & 0.9896 & 0.9890 \\
User tier & 0.9372 & 0.9363 & 0.9360 \\
Credit signup & 0.9485 & 0.9485 & 0.9427 \\
Credit limit & 0.8612 & 0.8613 & 0.8317 \\
Used credit & 0.9075 & 0.9077 & 0.8895 \\
Loan signup & 0.9510 & 0.9513 & 0.9478 \\
Account misuse & 0.7830 & 0.7835 & 0.7763 \\
Gambling & 0.9921 & 0.9921 & 0.9915 \\
Victim fraud & 0.8904 & 0.8906 & 0.8863 \\
Active fraud & 0.9551 & 0.9552 & 0.9526 \\
Automated abuse & 0.9269 & 0.9269 & 0.9224 \\
Crowd abuse & 0.8728 & 0.8732 & 0.8668 \\
Spending level & 0.9719 & 0.9716 & 0.9713 \\
Promo response & 0.7970 & 0.7966 & 0.7945 \\
\midrule
All-39 avg. & 0.8239 & 0.8239 & 0.8180 \\
Retention (\%) & 100.00 & 100.00 & 99.28 \\
\bottomrule
\end{tabular*}
\end{minipage}
\end{table*}


%% file: sections/04-experiments.tex
\section{Experiments}


{
We evaluate FastE's quality--efficiency trade-off, transfer across tasks, modalities, model scales and architectures, measured speedup, industrial applicability, and adaptive compression decisions.
The evaluation answers three research questions about our solution:}




\noindent
\textit{RQ1:} Can FastE preserve embedding quality while reducing inference cost across different settings?

\noindent
\textit{RQ2:} Can readout--prefix alignment determine when prefix-state compression should begin?

\noindent
\textit{RQ3:} Can readout-guided attention-score ranking determine which prefix states to retain?

{
Specifically, RQ1 corresponds to the quality, transfer, efficiency, and industrial results; RQ2 to the start-depth and threshold studies; and RQ3 to the ranking ablation.
}

\paragraph{Experimental Setup}


{
The evaluation covers five text-embedding benchmarks, three cross-modal retrieval tasks, two Qwen3-Embedding scales, and Qwen3-VL-Embedding.
Comparisons use matched target prefix-state budgets, and efficiency measurements include online triggering and selection overhead.
For both Qwen3 backbones, we use $l_{\mathrm w}=8$ and $\theta=0.60$, selected on the NarrativeQA validation split and then fixed across all test tasks and removal ratios.
The validation search and its quality--compute
criterion are described in the Hyperparameter Analysis below.
Appendix~\ref{app:evaluation-details} specifies further details about the tasks, baseline adaptations, compression settings, and timing procedures.
}

\begin{figure*}[t]
    \centering
    \includegraphics[width=0.98\textwidth]{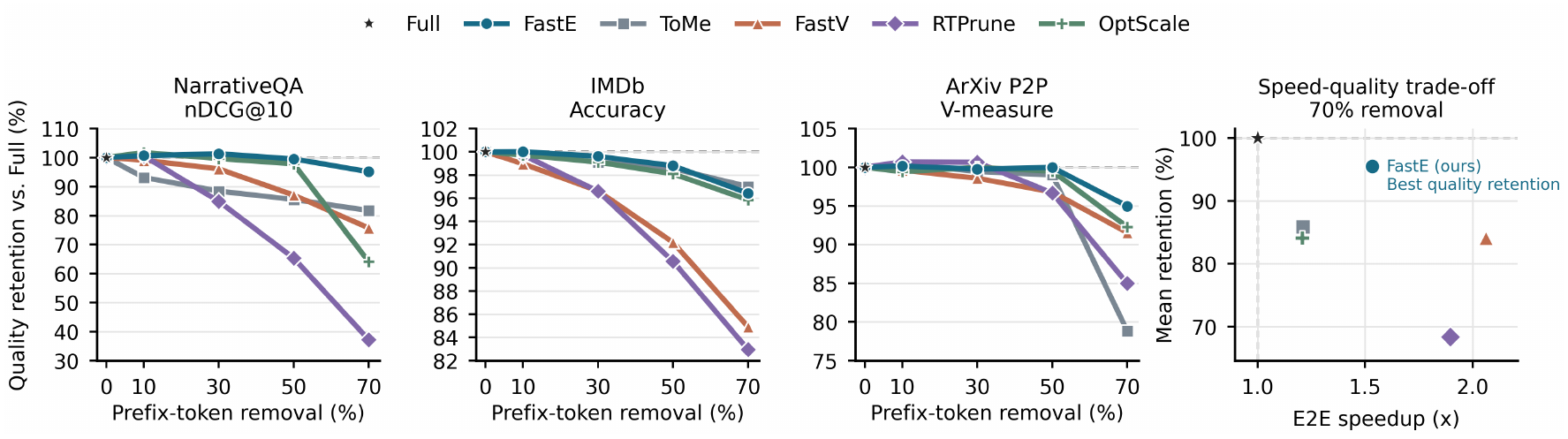}
    \caption{Quality retention across target removal ratios on NarrativeQA,
    IMDb, and ArXiv P2P using Qwen3-Embedding-0.6B. The first three panels show
    nDCG@10, Accuracy, and V-measure retention, respectively. The rightmost panel
    shows the quality--efficiency trade-off at 70\% removal, with mean quality
    retention across the three text tasks versus corpus end-to-end speedup. All methods use matched target
    per-sample prefix-state budgets; FastE uses $l_{\mathrm w}=8$ and $\theta=0.60$.}
    \label{fig:budget-sensitivity}
\end{figure*}

\begin{table}[t]
\centering
\scriptsize
\setlength{\tabcolsep}{1.4pt}
\caption{Cross-modal transfer on Qwen3-VL-Embedding-2B with
$l_{\mathrm w}=8$ and $\theta=0.70$. All tasks report Recall@10; T and I
denote text and image. Average retention is the three-task macro Recall@10
relative to Full Forward.}
\label{tab:cross-modal}
\begin{tabular*}{\columnwidth}{@{\extracolsep{\fill}}llcccc@{}}
\toprule
Dataset & Retrieval & Full & 30\% & 50\% & 70\% \\
\midrule
MSCOCO & T$\rightarrow$I & 0.9870 & 0.9870 & 0.9830 & 0.9520 \\
DocVQA & I+T$\rightarrow$T & 0.9970 & 0.9970 & 0.9970 & 0.9970 \\
NIGHTS & I$\rightarrow$I & 0.9930 & 0.9910 & 0.9830 & 0.9190 \\
\midrule
\multicolumn{2}{l}{Avg. retention $\uparrow$} & 100.00\% & 99.93\% & 99.53\% & 96.34\% \\
\bottomrule
\end{tabular*}
\end{table}

\subsection{Main Results}


{
Table~\ref{tab:main} compares downstream quality under matched target prefix-state budgets.
For each higher-is-better task score $s_t$, percentage retention is $R_t(\%)=100\,s_t/s_t^{\mathrm{Full}}$; mean retention averages $R_t$ over the five tasks, and mean retention loss is its complement to 100\%.
On Qwen3-Embedding-0.6B, FastE limits mean retention loss to 0.02\% at 30\% removal and 4.15\% at 70\%, compared with 2.53\% and 14.55\% for the strongest transferred baselines at the respective budgets.
The result persists on Qwen3-Embedding-4B, where FastE records 0.71\% and 3.07\% loss versus 2.36\% and 10.06\%.
Across both scales, Supply Chain Disclosure is comparatively stable, whereas retrieval and ArXiv clustering account for more of the residual loss.
}


{
\paragraph{Industrial benchmark.}
We evaluate FastE with Q-Anchor Embedding~\citep{yuan2026queryanchor} on 39 prediction tasks from proprietary Alipay industrial data. Non-industrial experiments use one A100; the industrial benchmark uses 40 L20 GPUs. For
readability, Table~\ref{tab:industrial-selected} reports selected tasks spanning
the major business categories. At 70\% prefix-state removal, the macro-average
AUC changes from 0.8239 to 0.8180, while the reported task-level AUCs remain
close to Full Forward. Complete task-level results are provided in
Appendix~\ref{app:industrial-results}.
}

{
\paragraph{Budget sensitivity.}
Figure~\ref{fig:budget-sensitivity} shows that FastE stays near Full Forward through 50\% removal on NarrativeQA, IMDb, and ArXiv P2P.
At 70\%, the NarrativeQA and ArXiv panels separate FastE most clearly from the transferred baselines, while IMDb remains less sensitive across methods.
The rightmost panel summarizes this behavior: FastE provides the highest mean quality retention, whereas faster alternatives obtain their speedups with substantially larger quality loss.
}

\paragraph{Cross-modal transfer.}

{
Table~\ref{tab:cross-modal} applies one shared $\theta=0.70$ rule to all serialized prefix positions in Qwen3-VL-Embedding.
At 50\% removal, MSCOCO, DocVQA, and NIGHTS each retain at least 98.99\% of Full Forward Recall@10, yielding 99.53\% average retention.
At 70\%, DocVQA remains unchanged and MSCOCO degrades modestly, but NIGHTS falls from 0.9930 to 0.9190 and reduces average retention to 96.34\%.
The contrast provides initial transfer evidence across natural-image, OCR-rich document, and visual-similarity retrieval while identifying 70\% removal as an aggressive cross-modal setting.
}

\paragraph{Cross-architecture transfer.}
Table~\ref{tab:e5-mistral-transfer} evaluates E5-Mistral-7B-Instruct on
NarrativeQA. FastE retains or slightly improves all three retrieval metrics
while achieving a $1.189\times$ total end-to-end speedup, extending the evidence
beyond the Qwen3 architecture family. Appendix~\ref{app:e5-mistral-transfer}
provides the interface, compression, and trigger details.

\begin{table}[t]
\centering
\scriptsize
\setlength{\tabcolsep}{2.0pt}
\caption{Cross-architecture transfer on NarrativeQA with
E5-Mistral-7B-Instruct. GPU-forward measures candidate-document encoding;
total E2E includes shared query encoding. Values are means over five paired runs.}
\label{tab:e5-mistral-transfer}
\begin{tabular}{lrrr}
\toprule
Metric & Full Forward & FastE & Change \\
\midrule
Recall@10 $\uparrow$ & 0.55929 & \textbf{0.56302} & +0.00373 \\
nDCG@10 $\uparrow$ & 0.45138 & \textbf{0.45477} & +0.00338 \\
MRR $\uparrow$ & 0.41745 & \textbf{0.42075} & +0.00330 \\
\midrule
Corpus E2E (s) $\downarrow$ & 178.30 & \textbf{130.22} & $1.369\times$ \\
GPU-forward (s) $\downarrow$ & 143.93 & \textbf{96.40} & $1.493\times$ \\
Total E2E (s) $\downarrow$ & 302.94 & \textbf{254.86} & $1.189\times$ \\
\bottomrule
\end{tabular}
\end{table}

\subsection{Real Inference Efficiency}

\begin{table}[t]
\centering
\scriptsize
\caption{NarrativeQA corpus-encoding efficiency across Qwen3-Embedding
backbones. Full denotes Full Forward.}
\label{tab:efficiency}
\setlength{\tabcolsep}{2.5pt}
\begin{tabular*}{\columnwidth}{@{\extracolsep{\fill}}llrrrr@{}}
\toprule
Model & Setting & nDCG@10 $\uparrow$ & FLOPs red. $\uparrow$
 & GPU-fwd. spd. $\uparrow$ & E2E spd. $\uparrow$ \\
\midrule
\multirow{4}{*}{0.6B} & Full & 0.45395 & 0\% & 1.000$\times$ & 1.000$\times$ \\
& FastE-30\% & \textbf{0.46005} & 26.19\% & 1.281$\times$ & 1.230$\times$ \\
& FastE-50\% & 0.45181 & 40.11\% & 1.540$\times$ & 1.363$\times$ \\
& FastE-70\% & 0.43179 & \textbf{51.19\%} & \textbf{1.934$\times$}
& \textbf{1.533$\times$} \\
\midrule
\multirow{4}{*}{4B} & Full & 0.58245 & 0\% & 1.000$\times$ & 1.000$\times$ \\
& FastE-30\% & \textbf{0.58005} & 28.05\% & 1.345$\times$ & 1.306$\times$ \\
& FastE-50\% & 0.57315 & 44.50\% & 1.694$\times$ & 1.606$\times$ \\
& FastE-70\% & 0.54637 & \textbf{59.15\%} & \textbf{2.309$\times$}
& \textbf{2.070$\times$} \\
\bottomrule
\end{tabular*}
\end{table}


{
Table~\ref{tab:efficiency} reports five-run speedups including alignment, ranking, and Top-$K$ overhead, with FLOPs computed from observed per-layer active sequence shapes.
For 0.6B, 50\% removal retains 99.53\% nDCG@10 while reducing FLOPs by 40.11\% and accelerating GPU-forward/E2E execution by 1.540$\times$/1.363$\times$; 70\% removal raises these gains to 51.19\% and 1.934$\times$/1.533$\times$ at 95.12\% retention.
The 4B rows show the same monotonic efficiency trend and a larger 70\% benefit: 59.15\% FLOPs reduction and 2.309$\times$/2.070$\times$ speedup.
Measured speedups remain below reciprocal-FLOPs estimates because they include online and kernel overhead.
}

\subsection{Ablation Studies}

\paragraph{Readout-triggered start depth.}

{
Table~\ref{tab:dynamic-vs-fixed} tests FastE's choice of when compression begins by comparing its readout--prefix alignment trigger with fixed layers L9, L11, L12, and L14.
All variants use the same 70\% target removal ratio, prefix-state budget, and attention-score ranking; Dynamic $\bar g$ uses $\theta=0.60$, triggers at mean layer 10.48, and is reported relative to Full Forward.
}

\begin{table}[t]
\centering
\scriptsize
\setlength{\tabcolsep}{1.6pt}
\caption{Start-depth ablation on NarrativeQA with Qwen3-Embedding-0.6B at
$r_{\max}=0.70$.}
\label{tab:dynamic-vs-fixed}
\begin{tabular*}{\columnwidth}{@{\extracolsep{\fill}}lcccc@{}}
\toprule
Start & nDCG@10 $\uparrow$ & Ret. $\uparrow$ & FLOPs red. $\uparrow$ & GPU-fwd./E2E $\uparrow$ \\
\midrule
Fixed L9 & 0.40351 & 88.89\% & 55.55\% & 2.082/1.641$\times$ \\
Fixed L11 & 0.41812 & 92.11\% & 49.70\% & 1.894/1.576$\times$ \\
Fixed L12 & 0.42898 & 94.50\% & 46.78\% & 1.692/1.476$\times$ \\
Fixed L14 & 0.41908 & 92.32\% & 40.93\% & 1.618/1.440$\times$ \\
\midrule
Dynamic $\bar g$ &
\textbf{0.43179} & \textbf{95.12\%} &
51.19\% & 1.934/1.533$\times$ \\
\bottomrule
\end{tabular*}
\end{table}


{
Fixed L9 saves the most FLOPs but filters prematurely, reducing nDCG@10 to 0.40351.
Dynamic $\bar g$ achieves the highest nDCG@10 (0.43179); relative to the strongest-quality fixed control, L12, it also improves FLOPs reduction from 46.78\% to 51.19\% and GPU-forward/E2E speedup from 1.692$\times$/1.476$\times$ to 1.934$\times$/1.533$\times$.
Later fixed L14 saves less compute without recovering quality, showing why a single predetermined depth is unreliable.
No fallback occurs in this evaluation; the rule avoids premature filtering rather than predicting an optimal layer, with supporting diagnostics in Appendices~\ref{app:alignment-distortion} and~\ref{app:alignment-commonality}.
}

\paragraph{Prefix-state ranking strategy.}
{
Table~\ref{tab:ablation} holds the Dynamic $\bar g$ trigger and target budget fixed to isolate the \textit{which} decision. Mean attention averages 32 sampled active queries, and random ranking averages seeds 42--44. Readout-guided, mean-attention, random, and position-only ranking score 0.43179, 0.42464, 0.1426, and 0.05006 nDCG@10, respectively.
}

{
Thus, Table~\ref{tab:dynamic-vs-fixed} supports the adaptive \textit{when} decision, and Table~\ref{tab:ablation} identifies the readout state as the informative signal for the \textit{which} decision.
}

\begin{table}[t]
\centering
\scriptsize
\setlength{\tabcolsep}{2.2pt}
\caption{Prefix-state ranking ablation on NarrativeQA with Qwen3-Embedding-0.6B at
$r_{\max}=0.70$.}
\label{tab:ablation}
\begin{tabular}{lccc}
\toprule
Variant & Start & Selection & nDCG@10 $\uparrow$ \\
\midrule
FastE & Dynamic $\bar g$ & Readout attn. & \textbf{0.43179} \\
w/o readout attn. & Dynamic $\bar g$ & Mean attn. & 0.42464 \\
w/o attention ranking & Dynamic $\bar g$ & Random & 0.1426 \\
Block-last & Dynamic $\bar g$ & Position-only & 0.05006 \\
\bottomrule
\end{tabular}
\end{table}


\paragraph{Hyperparameter analysis.}
\label{sec:hyperparameter-analysis}

\begin{figure}[t]
\centering
\includegraphics[width=0.78\columnwidth]{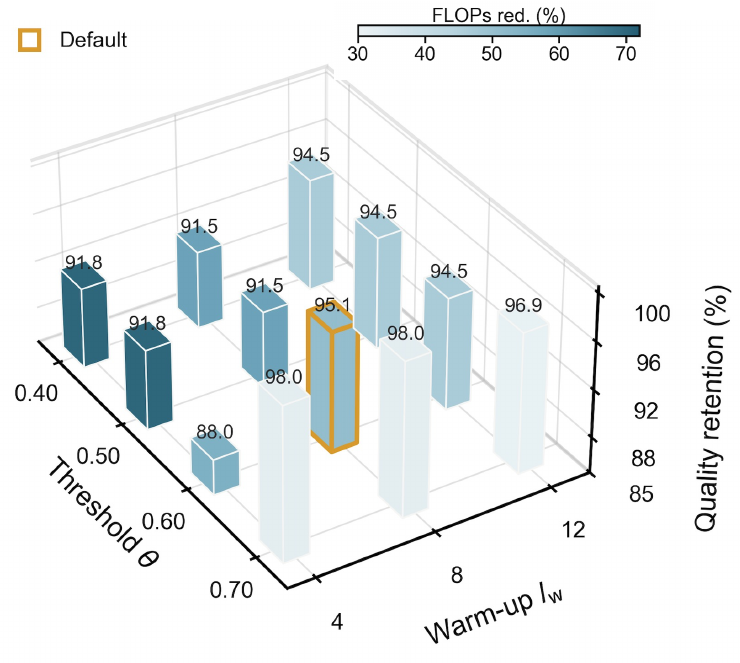}
\caption{Joint validation sensitivity of $l_{\mathrm w}$ and $\theta$ on
NarrativeQA with Qwen3-Embedding-0.6B at $r_{\max}=0.70$. Bar height denotes
validation quality retention and color denotes decoder-backbone FLOPs reduction.
The outline marks the selected configuration $(8,0.60)$.}
\label{fig:joint-trigger-sensitivity}
\end{figure}

Figure~\ref{fig:joint-trigger-sensitivity} summarizes the validation search
over $l_{\mathrm w}\in\{4,8,12\}$ and
$\theta\in\{0.40,0.50,0.60,0.70\}$ on NarrativeQA. Among configurations
retaining at least 95\% of Full Forward validation nDCG@10, $(8,0.60)$ provides
the largest decoder-backbone FLOPs reduction and is then fixed for all reported
test experiments. Lower thresholds trigger earlier compression, while larger
thresholds favor retention. Additional threshold, trigger-distribution, and
batch-size sensitivities are in Appendix~\ref{app:joint-trigger-sensitivity}.

%% file: sections/05-limitations.tex
\section{Limitations}

FastE targets final-readout LLM embedding models; final-layer pooling backbones
would require a pooling-aware compression signal. Its controller introduces
overhead, so FLOPs reductions should be interpreted with measured speedups.
The text evaluation and initial results on MSCOCO\_t2i, DocVQA, and NIGHTS
support the same modality-agnostic mechanism after serialization; broader
evaluation on video and additional multimodal backbones remains future work.

%% file: sections/06-conclusion.tex
\section{Conclusion}

FastE is a training-free method for reducing the cost of long-sequence
final-readout LLM embedding inference. Layer-wise results show that prefix-state
removal becomes better tolerated with depth. FastE uses batch-mean
readout--prefix alignment to trigger compression and readout attention to rank
the retained states, so later layers process a shortened sequence. On NarrativeQA
with Qwen3-Embedding-0.6B, it retains 99.53\% of Full Forward nDCG@10 with
40.11\% decoder-backbone FLOPs reduction and 1.363$\times$ end-to-end speedup.
Results across text tasks and three cross-modal retrieval tasks further support
the same final-readout interface for textual and visual prefixes, without
retraining or custom sparse-attention kernels.

%% file: appendices/01-additional-experimental-results.tex
\begin{center}
{\Large\bfseries Supplementary Material for}\\[3pt]
{\large\bfseries FastE: Readout-Triggered Token Compression for LLM Embedding Inference}
\end{center}
\vspace{0.75em}

\section{FastE Inference Procedure}
\label{app:faste-procedure}

Algorithm~\ref{alg:faste} summarizes the one-shot FastE inference path for one
batch. The full prefix is retained during warm-up and until the first threshold
crossing. If no crossing occurs earlier, the final-layer fallback
forces compression before the final transformer block.

The fallback was not activated in the Qwen3-Embedding-0.6B NarrativeQA threshold
sweep ($\theta\in\{0.50,0.60,0.70,0.80\}$) or in the E5-Mistral transfer run;
their trigger distributions are reported in the main text. It ensures a defined
compression path for a batch that never crosses the threshold.

\begin{algorithm}[!h]
\small
\caption{FastE inference for one batch}
\label{alg:faste}
\begin{algorithmic}[1]
\Require Batch $B$ with serialized states $\{H_i^0\}_{i\in B}$
\Require Blocks $\{F_l\}_{l=0}^{L-1}$; $l_{\mathrm w}$, $\theta$, and $r_{\max}$
\Ensure Sequence embeddings $\{z_i\}_{i\in B}$
\ForAll{$i\in B$}
    \State $R_i \gets \max\!\left(1,\operatorname{round}((1-r_{\max})N_i)\right)$
\EndFor
\State $\mathrm{compressed}\gets\mathrm{false}$
\For{$l=0,\ldots,L-1$}
    \If{$l\geq l_{\mathrm w}$ \textbf{and} $\neg\mathrm{compressed}$}
        \State Compute $g_i^l$ for every $i\in B$ using Eq.~\eqref{eq:sample-readiness}
        \State $\bar g_B^l \gets |B|^{-1}\sum_{i\in B}g_i^l$
        \If{$\bar g_B^l\geq\theta$ \textbf{or} $l=L-1$}
            \ForAll{$i\in B$}
                \State Compute $\mathbf w_i^l$ from attention received from the readout position
                \State $\mathcal S_i\gets\operatorname{sort}(\operatorname{TopK}(\mathbf w_i^l,R_i))$
                \State $H_i^l\gets\operatorname{Concat}([h_{i,j}^l]_{j\in\mathcal S_i},h_{i,r_i}^l)$
                \State Preserve the retained states' original rotary position IDs
            \EndFor
            \State $\mathrm{compressed}\gets\mathrm{true}$
        \EndIf
    \EndIf
    \State $H_i^{l+1}\gets F_l(H_i^l)$ for every $i\in B$
\EndFor
\State \Return $z_i\gets\operatorname{Norm}(h_{i,r_i}^L)$ for every $i\in B$
\end{algorithmic}
\end{algorithm}

\section{Additional Experimental Results}

\subsection{Additional Evaluation Details}
\label{app:evaluation-details}

\paragraph{Task protocols.}
For IMDb, we fit a logistic-regression probe on the training embeddings and
report test Accuracy. For each ArxivClusteringP2P.v2 subset, we sample at most
2,048 papers, fit MiniBatchKMeans using the gold number of categories, and
report mean V-measure. The cross-modal tasks represent text-to-image retrieval
(MSCOCO), visual-document retrieval (DocVQA), and image-to-image similarity
(NIGHTS).

\paragraph{Baseline adaptations.}
For a fair comparison, all baselines operate on the same serialized prefix,
use the same input truncation, batching protocol, and target removal ratios as
FastE, and always preserve the final readout and special tokens. ToMe performs
bipartite soft matching in each decoder block using cosine similarity between
the current-layer attention keys, merges matched states through a
cluster-size-weighted average, and applies proportional attention to account
for changes in cluster size. FastV follows its original one-shot pruning
setting: after the first two decoder blocks, prefix states are ranked by the
attention they receive, averaged over all valid query positions and attention
heads, and the lowest-scoring states are removed from all subsequent blocks.
RTPrune first selects dominant layer-0 prefix states according to their
$\ell_2$ norms, constructs an optimal-transport assignment from the remaining
states using pairwise cosine similarity, and merges their information into the
retained set. We use $z=0.2$, merge strength $\alpha=0.1$, and 100 Sinkhorn
iterations, while disabling its OCR-specific textual-density controller because
our inputs are serialized text rather than document images. OptScale
\citep{zhou2026statistically} adopts the same layer-wise matching and merging
schedule as ToMe and applies its parameter-free scaling rule to the merged value
vectors and attention logits according to the effective retention ratio at each
layer, thereby reducing representation-scale drift induced by repeated merging.
Fixed-layer filtering and random ranking serve as controls: they operate at
predetermined layers using either a fixed importance rule or uniform random
ranking and match FastE's final prefix-state budgets.

\paragraph{Timing protocol.}
End-to-end (E2E) timing includes tokenization, host-to-device transfer,
compression, and embedding output, whereas preloaded GPU-forward timing excludes
tokenization and data transfer. Both paths include alignment checks and the complete
trigger-layer scoring and Top-K selection. The main Qwen experiments
use length-sorted batches, batch size 4, and maximum length 5,000. Per-document
latency divides corpus time by the 355 NarrativeQA documents. Peak memory is
allocated CUDA memory; reserved allocator memory is diagnostic only.

\begin{table*}[!t]
\centering
\scriptsize
\setlength{\tabcolsep}{2.5pt}
\caption{Complete NarrativeQA efficiency records underlying
Table~\ref{tab:efficiency} for Qwen3-Embedding backbones. Full Forward is
reported once at 0\% removal for each backbone. GPU-forward and E2E entries give mean
seconds, with paired-run speedup in parentheses. Memory is peak allocated GiB.}
\label{tab:absolute-timing}
\renewcommand{\arraystretch}{1.05}
\begin{tabular}{@{}llrrrrrrr@{}}
\toprule
Backbone & Method & Removal
& Prefix-state rem.
& FLOPs red.
& nDCG@10
& GPU-fwd. (s / speedup)
& E2E (s / speedup)
& Mem. (GiB) \\
\midrule
\multirow{4}{*}{0.6B} & Full Forward & 0\% & 0\% & 0\% & 0.45395
& 48.57 (1.000$\times$) & 68.36 (1.000$\times$) & 3.875 \\
& FastE & 30\% & 30.00\% & 26.19\% & 0.46005
& 38.17 (1.281$\times$) & 56.38 (1.230$\times$) & 3.316 \\
& FastE & 50\% & 50.00\% & 40.11\% & 0.45181
& 31.09 (1.540$\times$) & 50.17 (1.363$\times$) & 2.948 \\
& FastE & 70\% & 69.98\% & 51.19\% & 0.43179
& 25.35 (1.934$\times$) & 44.02 (1.533$\times$) & 2.568 \\
\midrule
\multirow{4}{*}{4B} & Full Forward & 0\% & 0\% & 0\% & 0.58245
& 201.31 (1.000$\times$) & 221.12 (1.000$\times$) & 11.873 \\
& FastE & 30\% & 30.00\% & 28.05\% & 0.58005
& 149.57 (1.345$\times$) & 169.17 (1.306$\times$) & 10.817 \\
& FastE & 50\% & 50.00\% & 44.50\% & 0.57315
& 118.73 (1.694$\times$) & 137.12 (1.606$\times$) & 10.097 \\
& FastE & 70\% & 69.98\% & 59.15\% & 0.54637
& 87.17 (2.309$\times$) & 106.84 (2.070$\times$) & 9.697 \\
\bottomrule
\end{tabular}
\vspace{2pt}

\parbox{0.94\textwidth}{\scriptsize\emph{Note:} The 0.6B 70\% row uses the
updated five-run aggregate (25.35\,s GPU-forward; 44.02\,s E2E; 2.568\,GiB peak
allocated memory). Its paired-run speedups need not equal the ratio of the
displayed Full means. The 4B 30\% and 50\% speedups use separately paired Full
means (201.19/220.97\,s and 201.13/220.21\,s for GPU-forward/E2E, respectively); the
displayed 4B Full row reports the matched 70\% run.}
\end{table*}

\subsection{Joint Warm-Up and Threshold Sensitivity}
\label{app:joint-trigger-sensitivity}

The warm-up layer sets a lower bound on compression depth, while the alignment
threshold controls the quality--compute trade-off thereafter. Lower thresholds
generally trigger earlier and reduce more FLOPs, whereas $\theta=0.70$ favors
quality retention.
The validation-selected configuration $(l_{\mathrm w},\theta)=(8,0.60)$ retains
95.12\% of Full Forward nDCG@10 while reducing backbone FLOPs by 51.19\%.

\paragraph{Batch-size sensitivity.}
\begin{table}[t]
\centering
\scriptsize
\setlength{\tabcolsep}{2.4pt}
\caption{Corpus batch-size sensitivity on NarrativeQA with
Qwen3-Embedding-0.6B, $l_{\mathrm w}=8$, $\theta=0.60$, and
$r_{\max}=0.70$. Query embeddings use a fixed batch size of 4, while only the
length-sorted corpus batch size $B$ varies. Retention is relative to Full
Forward, and FLOPs are computed from observed per-layer active sequence
shapes.}
\label{tab:batch-size-sensitivity}
\begin{tabular*}{\columnwidth}{@{\extracolsep{\fill}}lrrrr@{}}
\toprule
$B$ & nDCG@10 $\uparrow$ & Ret. $\uparrow$ & Mean layer
& FLOPs red. $\uparrow$ \\
\midrule
1 & 0.43139 & 95.03\% & 10.41 & 51.41\% \\
2 & 0.43339 & 95.47\% & 10.46 & 51.26\% \\
4 (default) & 0.43179 & 95.12\% & 10.48 & 51.19\% \\
8 & 0.43602 & 96.05\% & 10.40 & 51.36\% \\
16 & 0.43637 & 96.13\% & 10.35 & 51.36\% \\
\bottomrule
\end{tabular*}
\end{table}

Across corpus batch sizes 1--16, nDCG@10 remains within
$[0.43139,0.43637]$, while the mean trigger layer varies by only 0.14 layers
and the analytical FLOPs reduction by 0.22 percentage points. Most batches
trigger at L10 or L11 under every setting. This result supports stability over
the tested length-sorted batch sizes, but does not establish invariance to
arbitrary batch composition.

\subsection{Threshold Sensitivity}
\label{app:threshold-sensitivity}

\begin{table}[t]
\centering
\scriptsize
\setlength{\tabcolsep}{2.4pt}
\caption{Threshold sensitivity on NarrativeQA with Qwen3-Embedding-0.6B at
$l_{\mathrm w}=8$ and $r_{\max}=0.70$.}
\label{tab:trigger}
\begin{tabular*}{\columnwidth}{@{\extracolsep{\fill}}lcccc@{}}
\toprule
$\theta$ & nDCG@10 $\uparrow$ & Ret. $\uparrow$ & Mean layer
& FLOPs red. $\uparrow$ \\
\midrule
0.50 & 0.41557 & 91.55\% & 8.00 & 58.47\% \\
0.60 & 0.43179 & 95.12\% & 10.48 & 51.19\% \\
0.70 & 0.44503 & 98.03\% & 16.88 & 32.53\% \\
0.80 & 0.45604 & 100.46\% & 22.00 & 17.54\% \\
\bottomrule
\end{tabular*}
\end{table}

Table~\ref{tab:trigger} shows that larger thresholds delay compression and
trade FLOPs savings for quality retention. The validation-selected default
$\theta=0.60$ retains 95.12\% of Full Forward nDCG@10 with 51.19\%
decoder-backbone FLOPs reduction.

\subsection{Representation-Space Discussion}
\label{app:alignment-commonality}

\begin{figure*}[!b]
    \centering
    \includegraphics[width=\textwidth,trim=0 0 0 7mm,clip]{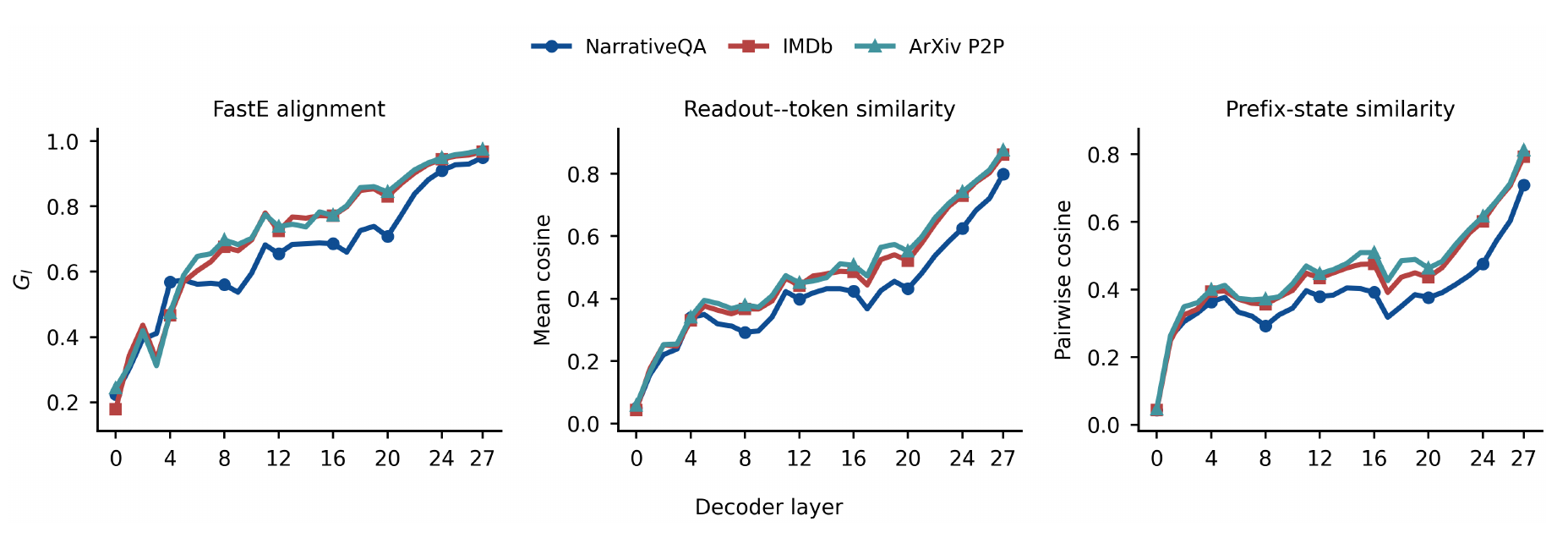}
    \caption{Representation-space diagnostics under uncompressed Full Forward
    inference with Qwen3-Embedding-0.6B. Lines report document-level means and
    shaded regions denote bootstrap 95\% confidence intervals. (a) FastE's
    readout--prefix alignment score $G$. (b) Mean cosine similarity between the
    readout state and individual prefix states. (c) Mean pairwise cosine similarity
    among prefix states.}
    \label{fig:alignment-commonality}
\end{figure*}

Figure~\ref{fig:alignment-commonality} examines why the alignment score increases
with depth. Across NarrativeQA, IMDb, and ArXiv P2P, the operational alignment score $G$
in panel (a), the mean readout--prefix similarity in panel (b), and prefix-state
commonality in panel (c) all increase toward deeper layers. The agreement
between panels (a) and (b) shows that the trend is not solely an artifact of
forming a raw prefix mean, while panel (c) indicates that the prefix states
themselves also become more homogeneous.

One plausible source of this functional redundancy is the combination of
sequence-level contrastive supervision at the final readout token and causal
information flow. Because the readout state is optimized as the sequence embedding
and can repeatedly attend to the full prefix, task-relevant information may
become increasingly concentrated in its representation. Meanwhile, causal
attention propagates earlier content into later prefix states, consistent with
the increasing representational overlap in panel (c). At greater depth, the
readout state may therefore have already absorbed more prefix information while the
remaining prefix states cover increasingly similar semantic directions,
reducing the marginal contribution of any individual prefix state.

Taken together, these three diagnostics support an interpretation in which the
readout state and increasingly homogeneous prefix states converge with depth. This
representation-space evidence complements, but does not replace, the controlled
compression interventions and alignment--distortion analysis below.

\subsection{Cross-Architecture Transfer on E5-Mistral}
\label{app:e5-mistral-transfer}

We evaluate FastE on E5-Mistral-7B-Instruct using its official interface by
appending EOS and extracting the last valid EOS state as the sequence embedding.
The NarrativeQA run uses batch size 2, maximum length 4,096,
$l_{\mathrm w}=8$, $\theta=0.40$, and $r_{\max}=0.70$. Query embeddings use
Full Forward and are shared by both conditions; FastE is applied only to the
355 candidate documents.
Table~\ref{tab:e5-mistral-transfer} reports the resulting quality and timing
comparison in the main paper. Table~\ref{tab:e5-threshold-sensitivity}
provides the corresponding local threshold sweep.

\begin{table}[t]
\centering
\scriptsize
\setlength{\tabcolsep}{3.4pt}
\caption{Post-hoc E5-Mistral threshold sensitivity on NarrativeQA with
$l_{\mathrm w}=8$ and $r_{\max}=0.70$. Retention is relative to Full Forward
(nDCG@10 $=0.45138$).}
\label{tab:e5-threshold-sensitivity}
\begin{tabular}{lrrr}
\toprule
$\theta$ & nDCG@10 $\uparrow$ & Ret. $\uparrow$ & Mean layer \\
\midrule
0.05 & 0.41033 & 90.90\% & 8.00 \\
0.10 & 0.41033 & 90.90\% & 8.00 \\
0.15 & 0.41661 & 92.30\% & 9.57 \\
0.20 & 0.46167 & 102.28\% & 17.60 \\
0.25 & 0.46196 & 102.34\% & 18.59 \\
0.35 & 0.45901 & 101.69\% & 20.28 \\
0.40 & 0.45477 & 100.75\% & 20.77 \\
0.50 & 0.45506 & 100.81\% & 21.76 \\
\bottomrule
\end{tabular}
\end{table}

FastE reduces the mean candidate-document sequence length from 4,096 to 1,230
hidden states, a 69.97\% realized prefix-state removal. It triggers at layers 20, 21, and
22 for 44, 131, and 3 batches, respectively, giving a mean trigger layer of
20.77 with no last-layer fallback. All three retrieval metrics remain at least
as high as Full Forward. Across five paired runs, mean corpus GPU-forward and
E2E speedups reach $1.493\times$ and $1.369\times$, respectively; mean total
E2E speedup, including shared query encoding, is $1.189\times$.

\subsection{Complete Industrial Task Results}
\label{app:industrial-results}

\paragraph{Protocol.}
Separately from the deployment-scale workload described in the Introduction,
we evaluate FastE on a controlled industrial task suite under a different
serving configuration. The evaluation uses the same proprietary Alipay
industrial data source as the Q-Anchor study and processes it with the published
Q-Anchor Embedding architecture~\citep{yuan2026queryanchor}, whose decoder
backbone is Qwen2.5-0.5B-Instruct. FastE uses
$l_{\mathrm w}=8$, $\theta=0.60$, and
$r_{\max}\in\{0.10,0.30,0.50,0.70,0.80\}$, with current-layer
readout-guided attention-score ranking and Top-K selection. The serialized user-information sequence has a
fixed length of 732 tokens. Inference is evaluated on 40 NVIDIA L20 GPUs with
batch size 256. The reported 2.4\,s and 1.0\,s values at 80\% removal are mean
GPU-forward times from the start of decoder processing of the serialized context
through construction of its prefix KV cache; input embedding, modality-specific
preprocessing, and subsequent prompt forwards are outside this interval. The
complete suite contains 39 prediction tasks spanning engagement, gaming,
payment, credit, risk-control,
anti-abuse, consumption, and preference scenarios. Table~\ref{tab:industrial-all}
reports task-level results and their macro-average; no raw user
records or identifying information are disclosed.

\begin{table*}[!t]
\centering
\scriptsize
\setlength{\tabcolsep}{1.8pt}
\renewcommand{\arraystretch}{0.95}
\caption{Complete AUC results on 39 Alipay industrial prediction tasks with Q-Anchor Embedding.
FastE-10\%, -30\%, -50\%, -70\%, and -80\% denote the prefix-state removal ratios. All settings
use the same downstream protocol. Macro avg. weights all tasks equally and is
computed from unrounded task scores; it can therefore differ by 0.0001 from an
average of the displayed rounded entries. Retention is relative to the Full
macro average.}
\label{tab:industrial-all}
\begin{tabular}{@{}llrrrrrr@{}}
\toprule
Category & Prediction task
& Full & FastE-10\% & FastE-30\% & FastE-50\% & FastE-70\% & FastE-80\% \\
\midrule
\multirow{3}{*}{Engagement} & Endpoint prediction & 0.8413 & 0.8413 & 0.8413 & 0.8413 & 0.8347 & 0.8353 \\
& RTA click & 0.6350 & 0.6350 & 0.6353 & 0.6352 & 0.6348 & 0.6351 \\
& MAU churn & 0.9703 & 0.9703 & 0.9703 & 0.9704 & 0.9683 & 0.9688 \\
\midrule
Ecosystem & Ecosystem activity & 0.9587 & 0.9587 & 0.9587 & 0.9587 & 0.9558 & 0.9567 \\
\midrule
\multirow{4}{*}{Gaming} & Gaming activity & 0.8985 & 0.8985 & 0.8986 & 0.8987 & 0.8807 & 0.8821 \\
& First game purchase & 0.9459 & 0.9459 & 0.9458 & 0.9457 & 0.9390 & 0.9386 \\
& Simulation-game category & 0.9495 & 0.9495 & 0.9493 & 0.9496 & 0.9467 & 0.9466 \\
& Board-game category & 0.8959 & 0.8959 & 0.8958 & 0.8961 & 0.8891 & 0.8897 \\
\midrule
\multirow{10}{*}{Payment} & Offline payment, next day & 0.7487 & 0.7487 & 0.7490 & 0.7494 & 0.7321 & 0.7322 \\
& Offline payment, seven-day segment & 0.7645 & 0.7645 & 0.7643 & 0.7644 & 0.7591 & 0.7582 \\
& Offline redemption, seven-day segment & 0.7358 & 0.7358 & 0.7352 & 0.7353 & 0.7312 & 0.7288 \\
& Channel-B payment, seven-day segment & 0.8178 & 0.8178 & 0.8178 & 0.8182 & 0.8127 & 0.8114 \\
& Channel-B payment, next day & 0.6960 & 0.6960 & 0.6961 & 0.6960 & 0.6817 & 0.6829 \\
& Channel-B redemption, seven-day segment & 0.7373 & 0.7373 & 0.7371 & 0.7374 & 0.7350 & 0.7320 \\
& Channel-B redemption, three-day segment & 0.6043 & 0.6043 & 0.6045 & 0.6042 & 0.6012 & 0.6017 \\
& Channel-B reward claim & 0.6567 & 0.6567 & 0.6572 & 0.6570 & 0.6444 & 0.6480 \\
& Channel-B post-scan conversion & 0.7380 & 0.7380 & 0.7380 & 0.7384 & 0.7282 & 0.7314 \\
& Payment-result-page click & 0.8903 & 0.8903 & 0.8909 & 0.8905 & 0.8880 & 0.8908 \\
\midrule
\multirow{3}{*}{Wealth} & Active-fund purchase & 0.6858 & 0.6858 & 0.6854 & 0.6870 & 0.6754 & 0.6776 \\
& Premium-card status & 0.9899 & 0.9899 & 0.9896 & 0.9899 & 0.9890 & 0.9848 \\
& User tier & 0.9372 & 0.9372 & 0.9363 & 0.9364 & 0.9360 & 0.9323 \\
\midrule
\multirow{4}{*}{Credit} & Credit-product signup & 0.9485 & 0.9485 & 0.9485 & 0.9486 & 0.9427 & 0.9361 \\
& Credit limit & 0.8612 & 0.8612 & 0.8613 & 0.8610 & 0.8317 & 0.8275 \\
& Used credit & 0.9075 & 0.9075 & 0.9077 & 0.9071 & 0.8895 & 0.8827 \\
& Loan-product signup & 0.9510 & 0.9510 & 0.9513 & 0.9511 & 0.9478 & 0.9473 \\
\midrule
\multirow{4}{*}{Risk control} & Account misuse & 0.7830 & 0.7830 & 0.7835 & 0.7834 & 0.7763 & 0.7761 \\
& Gambling & 0.9921 & 0.9921 & 0.9921 & 0.9921 & 0.9915 & 0.9915 \\
& Victim-fraud detection & 0.8904 & 0.8904 & 0.8906 & 0.8906 & 0.8863 & 0.8861 \\
& Active-fraud detection & 0.9551 & 0.9551 & 0.9552 & 0.9551 & 0.9526 & 0.9531 \\
\midrule
\multirow{2}{*}{Anti-abuse} & Automated abuse & 0.9269 & 0.9269 & 0.9269 & 0.9271 & 0.9224 & 0.9209 \\
& Crowdsourced abuse & 0.8728 & 0.8728 & 0.8732 & 0.8732 & 0.8668 & 0.8688 \\
\midrule
Consumption & Consumption capacity & 0.9719 & 0.9719 & 0.9716 & 0.9715 & 0.9713 & 0.9707 \\
\midrule
\multirow{7}{*}{Preference} & Brand sensitivity & 0.8281 & 0.8281 & 0.8274 & 0.8228 & 0.8216 & 0.8004 \\
& Promotion sensitivity & 0.7970 & 0.7970 & 0.7966 & 0.7945 & 0.7945 & 0.7883 \\
& Price sensitivity & 0.9058 & 0.9058 & 0.9046 & 0.9036 & 0.9002 & 0.8755 \\
& Achievement-reward preference & 0.5981 & 0.5981 & 0.5984 & 0.5976 & 0.5974 & 0.5950 \\
& Cash-reward preference & 0.6050 & 0.6050 & 0.6053 & 0.6052 & 0.6051 & 0.6035 \\
& Virtual-reward preference & 0.5874 & 0.5874 & 0.5875 & 0.5877 & 0.5879 & 0.5855 \\
& Physical-reward preference & 0.6552 & 0.6552 & 0.6551 & 0.6554 & 0.6542 & 0.6517 \\
\midrule
\multicolumn{2}{r}{Macro avg.}
& 0.8239 & 0.8239 & 0.8239 & 0.8238 & 0.8180 & 0.8160 \\
\multicolumn{2}{r}{Retention (\%)}
& 100.00 & 100.00 & 100.00 & 99.99 & 99.28 & 99.04 \\
\bottomrule
\end{tabular}
\end{table*}

\subsection{Decoder-Backbone FLOPs Accounting}
\label{app:flops-accounting}

We compute analytical decoder-backbone FLOPs from a shape-only pass over the
same length-sorted corpus batches used for efficiency measurement. For batch
$b$ at layer $l$, let $B_b$ be its batch size and $L_{b,l}$ its padded sequence
length. The realized padded-token and attention-pair counts are
\begin{equation}
    T_l=\sum_b B_bL_{b,l},
    \qquad
    P_l=\sum_b B_bL_{b,l}^2.
\end{equation}
Let $H$ and $I$ denote the hidden and SwiGLU intermediate dimensions, $A$ and
$A_{\mathrm{kv}}$ the numbers of query and key--value heads, and $d_h$ the head
dimension. Defining $Q=Ad_h$ and $K=A_{\mathrm{kv}}d_h$, and counting one
multiply--add as two FLOPs, the layer-wise components are
\begin{align}
    F_{\mathrm{proj}}^l
    &=2T_l\left[H(Q+2K)+QH\right],\\
    F_{\mathrm{attn}}^l
    &=4P_lAd_h,\\
    F_{\mathrm{MLP}}^l
    &=6T_lHI,\\
    \mathcal{C}_l(T_l,P_l)
    &=F_{\mathrm{proj}}^l+F_{\mathrm{attn}}^l+F_{\mathrm{MLP}}^l.
\end{align}
This accounting includes dense Q/K/V/O projections, dense QK/AV attention, and
the SwiGLU gate/up/down linear layers. It excludes embeddings, normalization,
RoPE, output pooling, alignment evaluation, trigger-layer scoring, and Top-K
selection; these auxiliary costs are included in measured GPU-forward and end-to-end
time.

For Qwen3-Embedding-4B on NarrativeQA at $r_{\max}=0.70$ and $\theta=0.60$,
separate Full Forward and FastE shape passes give
\begin{equation}
    1-\frac{C^{\mathrm{FastE}}}{C^{\mathrm{Full}}}
    =0.5914585.
\end{equation}
FastE therefore uses 40.85\% of Full Forward decoder-backbone FLOPs, which
corresponds to a FLOPs-implied speedup of approximately 2.45$\times$. The shape
passes do not repeat quality or timed inference; their purpose is only to record
the deterministic padded shapes entering each layer.

\subsection{Alignment--Distortion Diagnostic}
\label{app:alignment-distortion}

We examine how readout--prefix alignment relates to compression-induced
representation stability on the 89 NarrativeQA corpus batches used by the
$\theta=0.60$ experiment. Using the same 70\% target removal ratio and attention-score ranking as
the start-depth ablation, we intervene separately at every layer from L0 to L27.
Let $\tilde z_i^l$ be the final embedding when removal is applied at layer $l$,
and define the resulting distortion as
$D_i^l=1-\cos(z_i^{\mathrm{Full}},\tilde z_i^l)$. Across all monitored
batch--layer pairs, batch-mean alignment and distortion exhibit a strong pooled
Spearman association of $\rho=-0.942$: alignment generally increases across
network depth while compression-induced distortion decreases.

This depth-wise relationship is not perfectly synchronous. In particular,
alignment changes more slowly over the intermediate layers while mean
distortion continues to fall from 0.0240 at L10 to 0.00092 at L17. This pattern
is consistent with continued information integration after readout--prefix
alignment has begun to plateau; it does not establish that alignment alone
captures every aspect of representation refinement. After controlling for
layer as a categorical variable and for log sequence length, the partial
Spearman correlation is $\rho=-0.190$, with a 95\% bootstrap confidence interval
of $[-0.248,-0.125]$ over 2,000 batch-level resamples. The remaining conditional
association is modest but nonzero. We therefore interpret alignment primarily
as a depth-wise compression-readiness heuristic with limited batch-specific
information, rather than as an oracle for the accuracy-optimal layer of each
batch.

Operationally, FastE uses the first post-warm-up threshold crossing only to
gate the fixed compression schedule; it neither estimates a per-batch accuracy
optimum nor changes the prescribed removal budget.

\section{Transfer Configuration and Additional Results}
\label{app:transfer-additional-results}

\subsection{Backbone-Specific Settings}
\label{app:transfer-configurations}

\begin{figure}[t]
\centering
    \includegraphics[width=0.98\columnwidth]{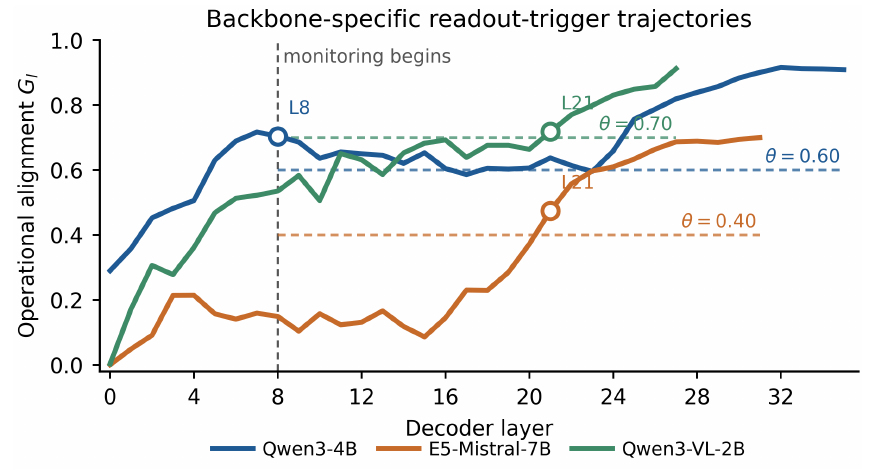}
\caption{Backbone-specific prefix-redundancy trajectories. $G_l$ denotes mean
readout--prefix alignment at decoder layer $l$. The vertical dashed line marks
the common warm-up depth $l_{\mathrm w}=8$. The trajectories show that prefix
redundancy emerges at backbone-specific rates.}
\label{fig:backbone-redundancy}
\end{figure}

Figure~\ref{fig:backbone-redundancy} shows that prefix redundancy emerges at
different rates across backbones. We therefore use backbone-local threshold
settings rather than transferring the text-Qwen threshold unchanged.
Table~\ref{tab:transfer-configurations} summarizes the settings used in the
reported cross-modal and cross-architecture experiments; the E5-Mistral sweep
above characterizes its local threshold range.

\begin{table}[t]
\centering
\scriptsize
\setlength{\tabcolsep}{3.0pt}
\caption{Configurations for reported transfer experiments.}
\label{tab:transfer-configurations}
\begin{tabular}{lccc}
\toprule
Backbone & $l_{\mathrm w}$ & $\theta$ & $r_{\max}$ \\
\midrule
Qwen3-VL-Embedding-2B & 8 & 0.70 & 0.30/0.50/0.70 \\
E5-Mistral-7B-Instruct & 8 & 0.40 & 0.70 \\
\bottomrule
\end{tabular}
\end{table}

\paragraph{Visual-prefix threshold rationale.}
Qwen3-VL uses a higher backbone-local threshold to avoid evicting visual-prefix
states before they receive sufficient intermediate-layer processing. This choice
is motivated by visual-specialized attention heads in the multimodal backbone,
whose contributions must be integrated into the final readout before
compression. Table~\ref{tab:vl-threshold-sensitivity} reports an auxiliary
Image-Long2 sweep on ChartQA and InfographicsVQA. Thresholds
$\theta\in[0.55,0.65]$ trigger at mean layers 9.4--12.4 and substantially
lower Recall@10 on both probes. Increasing the threshold to $\theta=0.70$
delays the mean trigger to L18.3 and substantially recovers quality. We
therefore hold $\theta=0.70$ fixed for the reported Qwen3-VL transfer
experiments. This sweep is a post-hoc diagnostic rather than a validation-based
hyperparameter selection and does not alter the fixed configuration of the
reported transfer experiments.

\noindent\begin{minipage}{\columnwidth}
\centering
\scriptsize
\setlength{\tabcolsep}{3.3pt}
\captionof{table}{Post-hoc Qwen3-VL threshold sensitivity on the Image-Long2 visual
probes with $l_{\mathrm w}=8$ and $r_{\max}=0.70$. All settings realize
69.93\% prefix-state removal.}
\label{tab:vl-threshold-sensitivity}
\begin{tabular}{rrrr}
\toprule
$\theta$ & ChartQA R@10 $\uparrow$ & InfoVQA R@10 $\uparrow$ & Mean layer \\
\midrule
0.55 & 0.703 & 0.679 & 9.41 \\
0.60 & 0.640 & 0.708 & 10.96 \\
0.65 & 0.661 & 0.692 & 12.42 \\
0.70 & 0.835 & 0.937 & 18.31 \\
0.75 & 0.872 & 0.942 & 22.08 \\
0.80 & 0.871 & 0.942 & 23.53 \\
\bottomrule
\end{tabular}
\end{minipage}

\paragraph{EOS-readout threshold rationale.}
E5-Mistral uses a lower backbone-local threshold because the raw alignment
scale of its EOS readout differs from that of Qwen. Table~\ref{tab:e5-threshold-sensitivity}
reports a post-hoc NarrativeQA sweep, rather than a held-out validation
selection. At $\theta\in[0.05,0.15]$, compression begins at mean layers
8.0--9.6 and nDCG@10 falls to 0.410--0.417, indicating premature eviction.
Over $\theta\in[0.35,0.50]$, nDCG@10 varies by only 0.00424 while the mean
trigger layer moves from 20.28 to 21.76. We use $\theta=0.40$ as an interior
operating point in this local range (mean trigger layer 20.77) and hold it
fixed throughout the E5 transfer experiment. Because $r_{\max}$ is fixed,
all settings realize the same 69.97\% prefix-state removal on candidate documents; this sweep
therefore isolates trigger timing and quality rather than the final token
budget.

\subsection{Additional E5-Mistral Results}
\label{app:e5-additional-results}

We additionally evaluate E5-Mistral on Core17 and two LegalBench groups using
the same fixed $l_{\mathrm w}=8$, $\theta=0.40$, and $r_{\max}=0.70$ setting.
Table~\ref{tab:e5-four-task-summary} summarizes the four completed tasks.
The metrics are task-specific, so retention is computed within each row and no
cross-task average is formed.

\begin{center}
\scriptsize
\setlength{\tabcolsep}{2.8pt}
\captionof{table}{Additional E5-Mistral transfer results with fixed
$l_{\mathrm w}=8$, $\theta=0.40$, and $r_{\max}=0.70$. Retention is relative
to Full Forward within each task. Corpus E2E speedup measures
candidate-document encoding.}
\label{tab:e5-four-task-summary}
\begin{tabular}{lrrrr}
\toprule
Task (metric) & Full $\uparrow$ & FastE $\uparrow$ & Ret. $\uparrow$ & Corpus E2E $\uparrow$ \\
\midrule
NarrativeQA (nDCG@10) & 0.45138 & 0.45477 & 100.75\% & 1.369$\times$ \\
Core17 (nDCG@10) & 0.46320 & 0.45065 & 97.29\% & 1.203$\times$ \\
Supply Chain (MacroAP) & 0.82238 & 0.83340 & 101.34\% & 1.222$\times$ \\
MAUD (MacroAccuracy) & 0.51073 & 0.50354 & 98.59\% & 1.106$\times$ \\
\bottomrule
\end{tabular}
\end{center}